\documentclass[10pt,journal,compsoc]{IEEEtran}
\usepackage{comment}
\usepackage{subfig}
\usepackage{graphicx} 
\usepackage{algorithm}
\usepackage{algorithmic}
\usepackage{bm}
\usepackage{amsmath}
 \usepackage{xcolor}
\usepackage{booktabs}

\usepackage{lipsum}
\usepackage{tabularx}

\ifCLASSOPTIONcompsoc
  \usepackage[nocompress]{cite}
\else
  \usepackage{cite}
\fi
\ifCLASSINFOpdf
\else
\fi
\begin{document}
%
\title{Improving Large Models with Small models: Lower Costs and Better Performance}
%
%
%
%

	\author{Dong~Chen, Shuo~Zhang, Yueting~Zhuang,~\IEEEmembership{Senior~Member,~IEEE}, Siliang~Tang, Qidong~Liu, Hua~Wang, Mingliang Xu,~\IEEEmembership{Member,~IEEE}   
\IEEEcompsocitemizethanks{\IEEEcompsocthanksitem D. Chen, Q. Liu, H. Wang and M. Xu are with the School of Computer and Artificial Intelligence of Zhengzhou University, and Zhengzhou University Engineering Research Center of Intelligent Swarm Systems, Zhengzhou, China (E-mail: chendongcs@zju.edu.cn; ieqdliu@zzu.edu.cn; iewanghua@zzu.edu.cn; iexumingliang@zzu.edu.cn).\protect\\
\IEEEcompsocthanksitem S. Zhang, Y. Zhuang and S. Tang are with Zhejiang University, Hangzhou, China (E-mail: shuo.zhang@zju.edu.cn; yzhuang@zju.edu.cn; siliang@zju.edu.cn).}
}

\IEEEtitleabstractindextext{%
\begin{abstract}
	Pretrained large models (PLMs), such as ChatGPT, have demonstrated remarkable performance across diverse tasks. However, the significant computational requirements of PLMs have discouraged most product teams from running or fine-tuning them. In such cases, to harness the exceptional performance of PLMs, one must rely on expensive APIs, thereby exacerbating the economic burden. Despite the overall inferior performance of small models, in specific distributions, they can achieve comparable or even superior results. Consequently, some input can be processed exclusively by small models. On the other hand, certain tasks can be broken down into multiple subtasks, some of which can be completed without powerful capabilities. Under these circumstances, small models can handle the simple subtasks, allowing large models to focus on challenging subtasks, thus improving the performance. We propose Data Shunt$^+$ (DS$^+$), a general paradigm for collaboration of small and large models. DS$^+$ not only substantially reduces the cost associated with querying large models but also effectively improves large models' performance. For instance, ChatGPT achieves an accuracy of $94.43\%$ on Amazon Product sentiment analysis, and DS$^+$ achieves an accuracy of $95.64\%$, while the cost has been reduced to only $31.18\%$. Besides, experiments also prove that the proposed collaborative-based paradigm can better inject specific task knowledge into PLMs compared to fine-tuning.
\end{abstract}

}

\maketitle

\IEEEdisplaynontitleabstractindextext

%
\IEEEpeerreviewmaketitle

\IEEEraisesectionheading{\section{Introduction}}
	\label{Introduction}
Recent years have seen a surge of interest in pretrained large models (PLMs) \cite{yu2023hallucidoctor,li2023gradient}, which are trained on a vast quantity of data at scale and can be adapted to a wide range of downstream tasks \cite{bommasani2021opportunities}. Large language models (LLMs), such as GPT \cite{brown2020language,ouyang2022training,liu2023pre} have demonstrated outstanding performance in text-related tasks \cite{wei2022emergent, brown2020language}. Additionally, multimodal large models \cite{zhu2023visual} like Flamingo \cite{alayrac2022flamingo} and BLIP-2 \cite{li2023blip} have been developed to extend the capabilities of LLM to encompass vision modality \cite{zhang2024vision,zeng2023x,xu2023multimodal}. PLMs, especially ChatGPT, have been extensively applied across diverse domains such as coding \cite{surameery2023use}, education \cite{biswas2023role}, and health \cite{biswas2023role}, fundamentally transforming people’s lives.

\begin{figure}[t]
	\centering
	\includegraphics[scale=0.36]{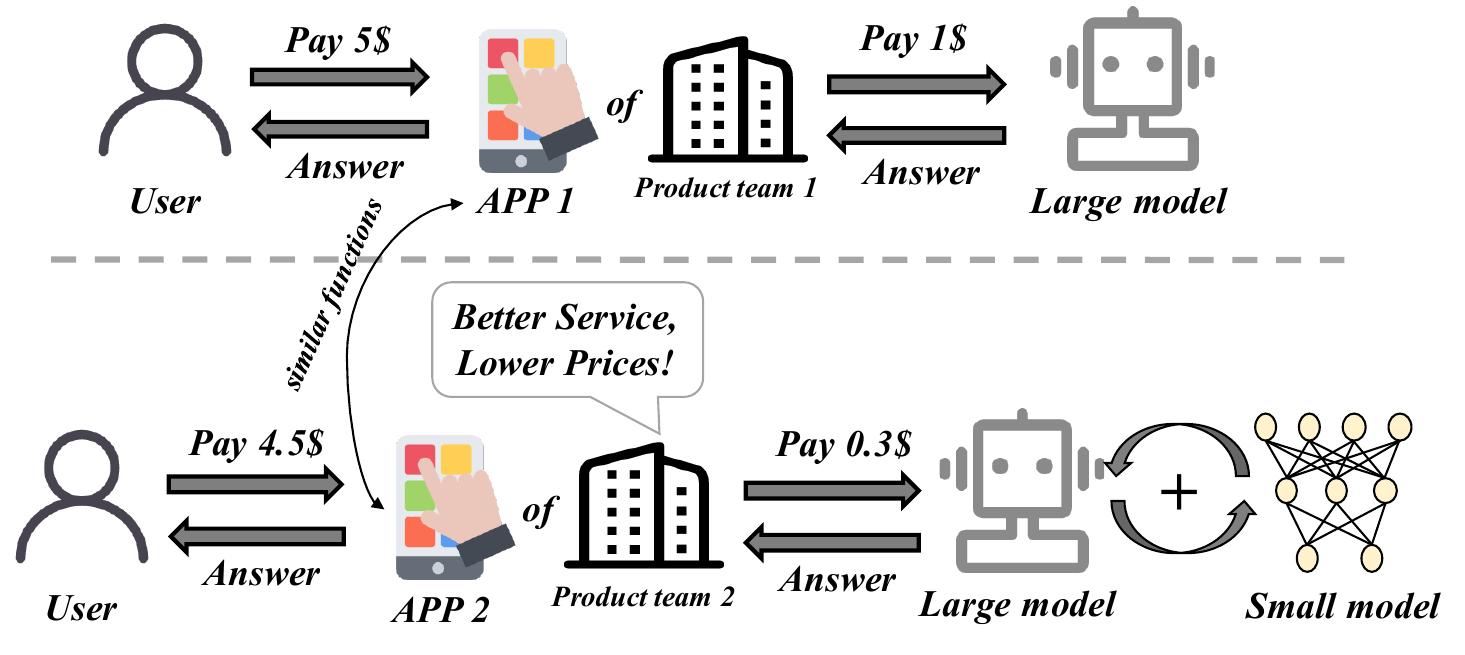}
	\caption{Commercial Applications of Large Models. \textit{upper: } Product team 1 only use large models to support their applications. \textit{lower: } Product team 2 reduces costs by collaborating with both large and small models, thereby allowing them to provide more appealing prices to users. }
	\label{company_instance}
\end{figure}

Despite the impressive performance of PLMs across various applications, their computational requirements render them impractical for deployment on numerous devices. As a result, product teams might opt to acquire the interface of the PLMs to access the associated services. Nonetheless, frequent invocations of the interface prove to be prohibitively costly for product teams.
Although small models' overall performance is significantly lower than that of large models, it can still outperform or achieve highly competitive results across certain data distributions.
Based on the above idea, we first divide samples into easy samples and hard samples. 
Easy samples represent the majority of the training data and are relatively easier for small models to learn and predict accurately. In contrast, hard samples refer to data that poses challenges for small models, including samples that deviate from the main distributions of the training data and samples located at the boundaries between different categories. 
For example, for long-tail data with few majority classes (head) and large amount of minority classes (tail) \cite{ouyang2016factors,zhang2017range,oksuz2020imbalance}, small models can perform much better on head data, thus, head data is easy samples, while tail data is hard samples. Moreover, small models often succumbs to overfitting when trained on a limited dataset. During inferencing, small models demonstrate good performance when the inputs resemble the training data. However, small models exhibit poor performance when the inputs significantly deviates from the training data \cite{horenko2020scalable}. In such scenarios, data that conforms to the training distribution is considered as easy samples, while data that deviates from the training distribution is considered as hard samples.
For PLMs, there is no distinction between easy and hard samples since these models have already been exposed to substantial volumes of training data. The distribution of unknown samples is unlikely to deviate significantly from the knowledge acquired by the large models. This is also the reason why PLMs exhibit significantly better performance in real-world applications. For convenience, we designate powerful yet challenging-to-deploy PLMs as \textbf{large models} and models that are easy to train and deploy as \textbf{small models}. We propose to improve the performance of large models by integrating specialized knowledge from specific small models, concurrently reducing the querying costs associated with large models.

As illustrated in Fig. \ref{company_instance}, APP 1 and APP 2 have similar functions, and APP 1 only uses large models, incurring additional costs compared to APP 2 that involves collaboration between large and small models. By reducing costs, APP 2 can offer more appealing prices, thereby enhancing market competitiveness.
In this work, we mainly focus on how to improve the performance of large models with specific small models, while reducing the cost of querying large models.
Specifically, we propose Data Shunt$^+$ (DS$^+$), which utilizes the confidence of small models to determine the appropriate processing direction for the input data: either through large models or exclusively through small models. If the confidence level exceeds a threshold, the input will be processed by small models; otherwise, large models will be involved in handling the input.
DS$^+$ is a collaborative paradigm involving both small models assisting large models (Small Model for Large Model, S4L) and large models assisting small models (Large Model for Small Model, L4S).
For S4L, we propose Prompt Pruning (PP) and Prompt Transferring (PT). PP aims to refine the prediction space of large models with the help of small models. In a classification task, as a sample transitions from the small model to the large model, the probability of it belonging to distributions in which the small model demonstrates proficiency diminishes. Thus, this probability can be integrated into the prompt to enhance the large model’s discrimination against alternative distributions.
On the other hand, PT attempts to breaks down a complex task into multiple sub-tasks, among which are simple tasks that can be processed by small models. By decomposing the task and allowing small models to manage certain intermediate steps, large models can focus on difficult sub-tasks, thereby improving the performance. Simultaneously, processing certain sub-tasks with small models usually leads to an effective decrease in the number of tokens in the input for large models, thus reducing the cost.
As for L4S, large models, equipped with extensive general knowledge, can distill less common knowledge to improve small models and further reduce the reliance on large models. However, the distillation process often leads to catastrophic forgetting since small models are initially proficient only in a limited number of distributions. To address this issue, we introduce 2-Stage Confidence Distillation (2CD), a method where small models learn iteratively from high-confidence samples provided by large models and their original version.


The effectiveness of the proposed method is validated across various modalities and tasks. Specifically, in the sentiment analysis task, DS$^+$ enhances the overall accuracy of ChatGPT by $1.21\%$, while reducing the cost of the large model to $31.18\%$ of its original expense. Besides, DS$^+$ improves the performance of ChatGPT on the dataset with inherent ambiguities from $63.73\%$ to $67.29\%$, while the cost is reduced to $58.54\%$ of the original.
In the image classification task, DS$^+$ improves the overall accuracy by $5.07\%$, and the number of invocations of the large model has been reduced to $66.10\%$ of its original frequency. Additionally, for the image caption task, DS$^+$ elevates the average BLEU score by 0.42, with the number of invocations reduces to $65.36\%$.

The main contributions of this paper can be summarized as follows:
\begin{itemize}
	\item We introduce a new paradigm to improve the performance of large models while reducing costs, which involves the collaboration between small models with specialized knowledge and large models with general knowledge.
	\item  We introduce Small Model for Large Model (S4L) and Large Model for Small Model (L4S), which further improve the performance while reducing costs.
	\item We compare the proposed method with fine-tuning, demonstrating the effectiveness of collaborative-based paradigm and providing a new perspective for the injection of specific knowledge.
	\item We demonstrate the efficacy of the proposed method across diverse multimodalities and tasks.

\end{itemize}

\section{Related Work}


\subsection{Large Model and Small Model}

Deploying large models with remarkable few-shot capabilities \cite{smith2022using,zhang2022opt,hoffmann2022training,li2023finetuning} poses a challenge in real-world applications primarily due to their enormous size.
For example, running a 175 billion LLM requires at least 350GB GPU memory \cite{zheng2022alpa}, which are far beyond affordable for most product teams, let alone more large models over 500B parameters \cite{chowdhery2022palm}. 
Moreover, the costly interface also poses challenges in addressing real-world issues using large models, as its expensive cost exceeding the affordability of most product teams.
Consequently, we propose to mitigate the aforementioned issues by employing smaller, specialized models. In addition, there are also studies focus on large and small models.

\subsection{Reducing Cost and Improving Performance}
Prior works discuss three main strategies for cost reduction: prompt adaptation, LLM approximation, and
LLM cascade \cite{chen2023frugalgpt}. 
The prompt adaptation try to make the prompt shorter.
LLM approximation explores how to create simpler and cheaper LLMs on specific tasks. 
LLM cascade aims to adaptively choose different APIs for
different queries. Different from prior works that only focus on language modality, we explores a generalized paradigm that can be applied to various modalities and tasks. Besides, we reduce large model cost by combining task-specific small models, which is simple yet effective.

\section{Methodology}

The proposed Data Shunt$^+$ (DS$^+$) is a collaborative paradigm of large and small models based on Data Shunt \cite{chen2024data}. 
In the subsequent sections, we will explore how small models contribute to the improvement of large models, how large models benefit small models, as well as provide an overview of the entire DS$^+$ paradigm.

\subsection{Small Models for Large Models}
\begin{figure}[t]
	\centering
	\includegraphics[scale=0.35]{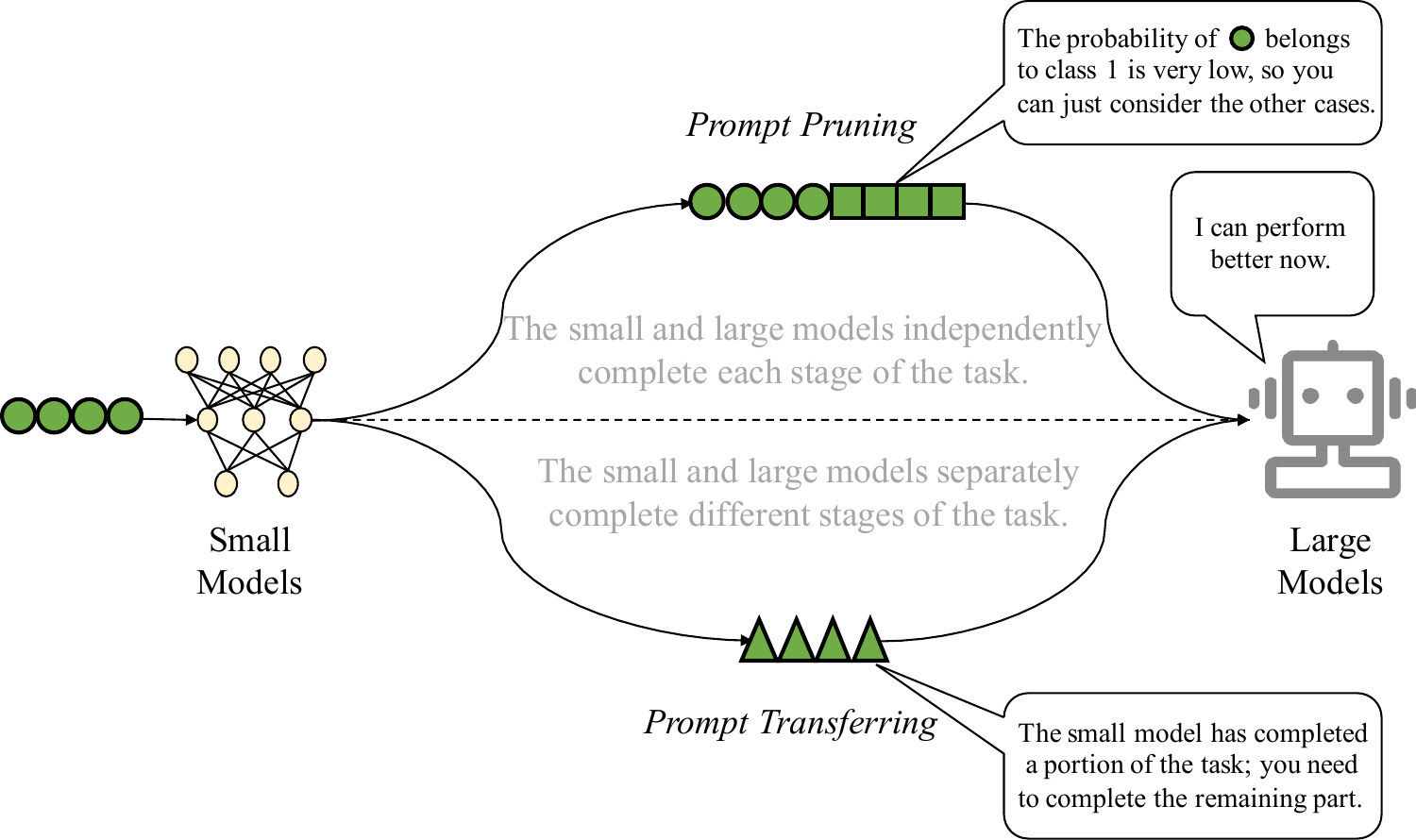}
	\caption{Small Model for Large Model (S4L). There are two methods in S4L, including Prompt Pruning (PP) and Prompt Transferring (PT). PP refines the prediction space of large models, while PT refines the input space of large models.}
	\label{S4L}
\end{figure}

As shown in Fig. \ref{S4L}, we focus on Small Model for Large Model (S4L) in this section. 

Small models can exhibit superior performance when dealing with specific data distributions. To improve the performance of large models with advantages of small models, we propose Prompt Pruning (PP) for classification task. PP refines the prediction space of large models using the prompts crafted with predictions of small models.

As demonstrated in Equation \ref{Small_model}, we obtain the prediction confidence $C_s$ by subjecting the output of a trained small model $F_{s}$ to a softmax operation. A higher value in a specific dimension of $C_s$ indicates a greater level of confidence from the small model regarding the corresponding judgment.
\begin{equation}
\begin{aligned}
C_{s} = \frac{e^{z_i}}{\sum e^{z_d}},\quad z_i \in F_{s}(x)
\label{Small_model}
\end{aligned}
\end{equation}
where $x$ is the input data.

Intuitively, small models excel at discriminating specific distributions. For instance, a small model may excel at distinguishing cats from other animals. Although it cannot recognize dogs, tigers, and other animals, it can confidently determine that they are not cats (i.e., the corresponding confidence is lower). Therefore, incorporating these predictions into the prompts for large models can refine the prediction space and enhance the performance of large models. 
For example, a prompt of PP for image classification task:

\textit{"This is a photo of a {label} with probability {$C_{s}$}."}

Compared to traditional prompt, PP introduces the confidence of small models as prior knowledge. 
When the input data do not follow distributions that small models excel at, $C_{s}$ of these distributions (or called classes) will be lower. If there are numerous candidate classes, we only add probability to classes small models excel at. Thus, large models can focus on other candidate classes and improve performance.
We call such prompt that with small models' confidence as soft prompt.

On the other hand, we can directly remove classes small models excel at. Thus, the prediction space is smaller, and the possibility for large models to predict the correct class will be higher. We call such prompt that directly remove the candidates as hard prompt. We further perform theoretical analysis from an entropy perspective to show that PP with soft and hard prompt can effectively improve the performance of large models. 

Let ${X}$ and ${Y}$ be the variable of input data and small model prediction, respectively. We use entropy to quantify the lower bound of model capability, where higher entropy indicates that the model struggles to produce effective results. $H({X})$ is the entropy of the input data and $H({Y})$ is the entropy of the prediction. We first show the effectiveness of soft prompt.
We use $H({X}\mid {Y})$ to represent the entropy of the input data with soft prompt $\hat{{X}} ={X}\mid {Y}$. 
Now we show that with ${Y}$, $H(\hat{{X}})$ is lower than $H({X})$:
\begin{equation}
\begin{aligned}
&H(X)-H(\hat{X})=
\sum_{x \in {X}} \sum_{y \in {Y}} p(x, y) \log _2 \frac{p(x, y)}{p(x) p(y)}\\
&\geq\left[\sum_{x \in {X}} \sum_{y \in {Y}} p(x, y)\right] \log _2 \frac{\sum_{x \in {X}} \sum_{y \in {Y}} p(x, y)}{\sum_{x \in {X}} \sum_{y \in {Y}} p(x) p(y)}=0
\end{aligned}
\end{equation}
Note that the above inequality takes the equal sign iff ${X}$ and ${Y}$ are independent, i.e., $p(x,y)=p(x)p(y)$. However, small models often exhibit a strong correlation between their outputs and inputs.
Thus, we get $H({X}) > H(\hat{{X}})$, which validates that the input data with soft prompt gets lower entropy. 

As for the hard prompt,  let entropy of the prediction of large models be $H({C_l})=-\sum_{i=1}^N c_i \log c_i$, where $N$ is the number of candidates, $\sum_{i=1}^N c_i-1=0$ and $c_i \in C_l$. With Lagrange multiplier method, we get:

\begin{equation}
G\left(c_1, c_{2 \ldots} c_N, \lambda\right)=-\sum_{i=1}^N c_i \log c_i+\lambda\left(\sum_{i=1}^N c_i-1\right)
\label{Lagrange}
\end{equation}
then partially differentiating $G$ in Equation \ref{Lagrange} with respect
to $c_i$ and $\lambda$,

\begin{align}
\begin{aligned}
\frac{\partial G}{\partial c_i}=-\log c_i-1+\lambda, \qquad
\frac{\partial G}{\partial \lambda}=\sum_{i=1}^N c_i-1
\end{aligned}
\label{KKT}
\end{align}

Let Equation \ref{KKT} be 0, we can get $c_1=c_2=\ldots=c_N=\frac{1}{N}$, $H({C_l})=\log N$, which is the maximum value of $H({C_l})$. If we perform PP with hard prompt, the number of candidates will be $M$, and $M<N$. Thus, the maximum value of $H({C_l})$ will be $\log M$. As $\log M < \log N$, the maximum value of $H({C_l})$ will be smaller, and the lower bound of large models will be higher. We defer the proof to the \textbf{\textit{Appendix A}}.

As for Prompt Transferring (PT), a complex task will be decomposed into multiple sub-tasks, with small models responsible for the relatively simpler sub-tasks. The output of small models will be combined with prompts and input into large models to complete difficult tasks. The processing can be represented as follows:

\begin{equation}
\begin{aligned}
z_{L} = F_{L}(z_i,prompt),\quad z_i \in F_{s}(x)
\label{EQ:PT}
\end{aligned}
\end{equation}
where $F_{L}$ is the function of large models, $z_{L}$ is the output of large models.

\subsection{Large Models for Small Models}
\begin{figure}[t]
	\centering
	\includegraphics[scale=0.47]{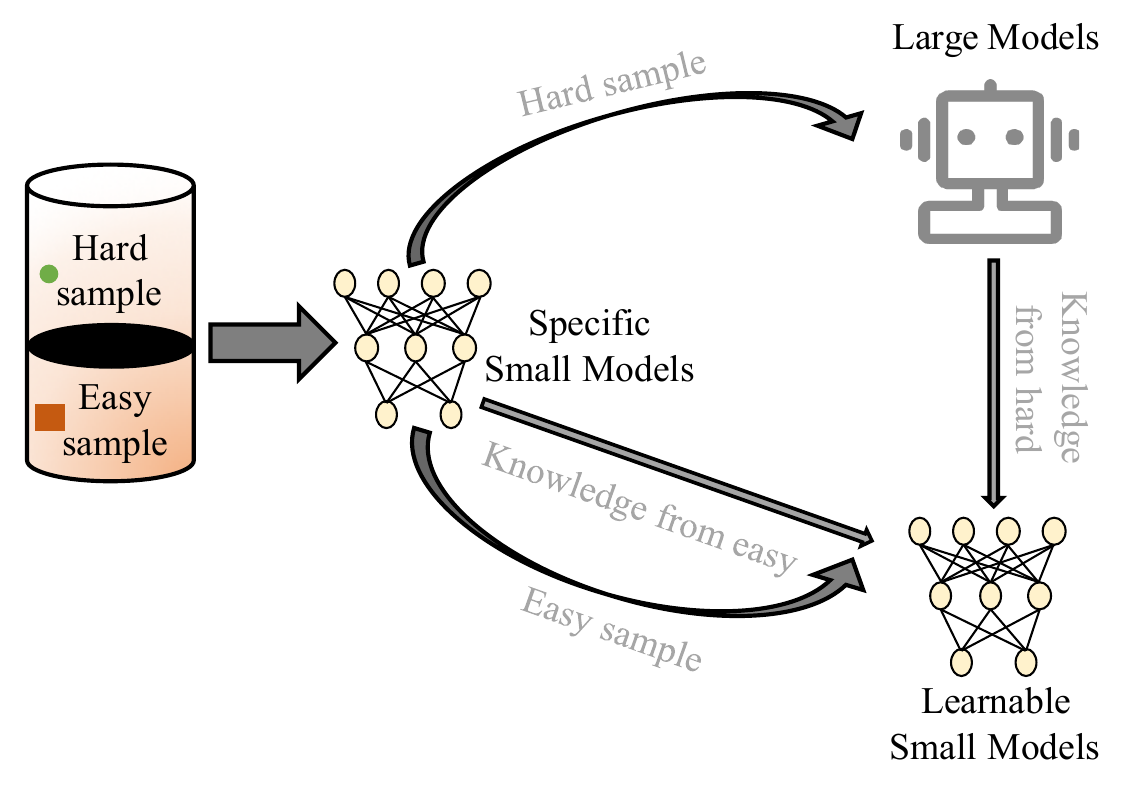}
	\caption{Large Model for Small Model (L4S). By injecting the knowledge of large models into small models, more samples can be transformed into easy samples, thereby further reducing the frequency of querying large models.}
	\label{L4S}
\end{figure}
In addition to small models assisting large models, we can also employ large models to support small models, enabling the distillation of knowledge that small models lack. As the knowledge of small models expands, they can handle a greater number of samples, thereby further reducing the frequency to query large models.
However, we observed that small models tend to forget the original well-fitting distributions after knowledge distillation \cite{french1999catastrophic,gou2021knowledge}. Besides, large models may severely degrade small models if the performance of large models is not well (i.e., distill incorrect knowledge). 
In order to address this issue, we propose 2-Stage Confidence Distillation (2CD), which performs knowledge distillation based on the confidence levels of both small and large models.
As illustrated in Fig. \ref{L4S}, to preserve the advantages of small models, we maintain a version of the original small models that do not receive knowledge from large models (referred to as specific small models).
Besides, we duplicate the specific small models to enhance knowledge acquisition (referred to as learnable small models). In order to mitigate the negative impact of incorrect knowledge on the learnable small models, we enable them to learn from both specific small models and large models simultaneously. 
For an input sample, when the confidence of specific small models is low and the confidence of large models is high, learnable small models will acquire predictions from the large models to introduce additional knowledge. Such process enables learnable small models to handle increasingly diverse samples. Conversely, learnable small models will continue to learn high confidence samples from specific small models to mitigate the impact of distorted knowledge from large models and prevent the forgetting of knowledge acquired during the training phase.

For an input data $x$, if $C_{s1}$ (computed by Equation \ref{Small_model}) is lower than shunt threshold, $\delta$, we compute the prediction of large models by
\begin{equation}
\begin{aligned}
C_{l} = \frac{e^{z_i}}{\sum e^{z_d}},\quad z_i \in F_{l}(x)
\label{Large_model}
\end{aligned}
\end{equation}
where $F_{l}$ is the function of large models and $C_{l}$ is the predicted confidence.

If $C_{l}>\delta$, we perform knowledge distillation with Kullback Leibler divergence \cite{hershey2007approximating} by 
\begin{equation}
\begin{aligned}
L_{ls} = KL(F_{s2}(x),C_{l})
\label{KL_ls}
\end{aligned}
\end{equation}	
To alleviate the impact of distorted knowledge from large models, we also select samples that $C_{s1}>\delta$ to perform knowledge distillation, where

\begin{equation}
\begin{aligned}
L_{s1s2} = KL(F_{s2}(x),C_{s1})
\label{KL_ss}
\end{aligned}
\end{equation}	

During 2CD, we perform Equation \ref{KL_ls} and \ref{KL_ss} iteratively.

\subsection{Data Shunt$^+$}
\begin{figure*}[t]
	\centering
	\includegraphics[scale=0.5]{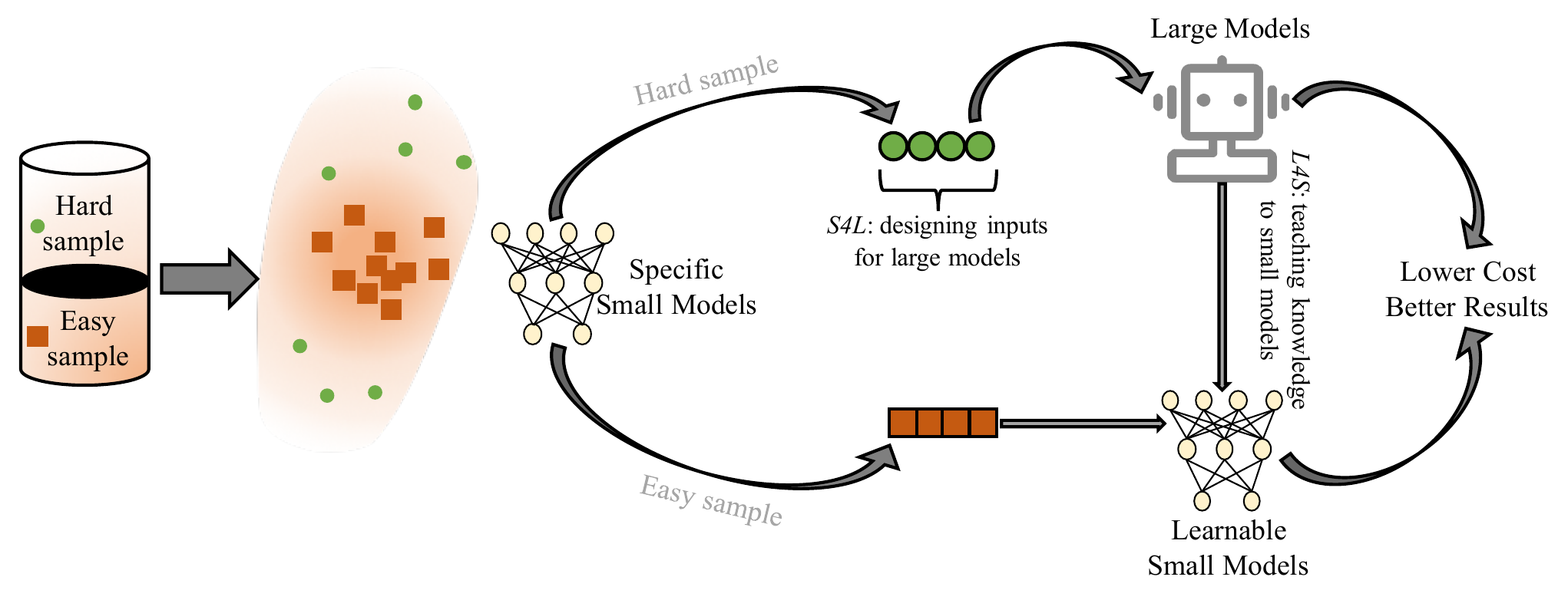}
	\caption{The training process of the proposed method. Hard samples refer to data that poses challenges for small models, while easy samples represent data that small models can fit well.}
	\label{DS}
\end{figure*}
We propose a collaborative paradigm, DS$^+$, which aims to mitigate the substantial cost associated with large models while simultaneously improving performance. 
During training, as illustrated in Fig. \ref{DS}, DS$^+$ tries to achieve a coordinated state between small and large models. On one hand, it performs knowledge distillation for the learnable small models, transforming more data into easy samples that the learnable small models can handle. On the other hand, DS$^+$ determines the shunt threshold $\delta$ by evaluating the specific small models’ confidence with training set, thus allowing more easy samples to be processed by the small models.
During inference, an input will be processed by specific small models and learnable small models at first. If the confidence of small models is lower than $\delta$, this sample and the corresponding confidence will be sent to large models, where the prompt will be re-designed with the confidence of small models.
On the other hand, if the confidence of the input sample is high, this sample will only be processed by small models.
The training process and inference process is summarized in Algorithm \ref{DS_training} and \ref{DS_inference}.

\begin{algorithm}[!h]
	\caption{Training of Data Shunt$^+$}
	\begin{algorithmic}[1]
		\STATE For an input data $x$; specific small models, $F_{s1}$; large models, $F_l$; shunt threshold $\delta$.
		\STATE Copy $F_{s1}$ and get learnable small models, $F_{s2}$.
		\STATE Compute the confidence $C_{s1}$ by Equation \ref{Small_model} and $F_{s1}$.
		\STATE Select data that $C_{s1} > \delta$ as $x_1$.
		\STATE Select data that $C_{s1} < \delta$ as $x_2$.
		\STATE Compute the confidence $C_{l}$ by Equation \ref{Large_model} and $F_{l}$ on $x_2$.
		\STATE Select data that $C_{l} >\delta$ from $x_2$ as $x_3$.
		\STATE Optimize $F_{s2}$ by $KL(F_{s2}(x_3),C_{l})$ and $KL(F_{s2}(x_1),C_{s1})$, iteratively.
	\end{algorithmic}
	\label{DS_training}
\end{algorithm}

\begin{algorithm}[!h]
	\caption{Inference of Data Shunt$^+$}
	\begin{algorithmic}[1]
		\STATE For an input data $x$, specific small models, $F_{s1}$, large models $F_l$, shunt threshold $\delta$.
		\STATE Compute the confidence $C_{s1}$ by Equation 1 and $F_{s1}$.
		\IF{$C_{s1}>\delta$}
		\STATE The prediction is generated by specific small models.
		\ELSE 
		\IF{$C_{s2}>\delta$}
		\STATE The prediction is generated by learnable small models.
		\ELSE 
		\STATE \textbf{\textit{PP:}} Generate prompt by PP, and the prediction is generated by large models.
		\STATE \textit{\textbf{PT:}} Decompose the task, and calculate the final result by Equation \ref{EQ:PT}.
		\ENDIF 
		\ENDIF 
	\end{algorithmic}
	\label{DS_inference}
\end{algorithm}

\begin{table*}[t]
				\caption{Different small models for DS$^+$ on sentiment analysis. Small model 1 represents TextCNN, small model 2 represents LSTM, and small model 3 represents fine-tuned BERT. The large mode is ChatGPT. For DS$^+$ we present the accuracy and query proportion (i.e., sample proportion processed by ChatGPT) in the same unit.}
\label{DS_on_AS_}
	\centering
\begin{tabularx}{\linewidth}{cX<{\centering}X<{\centering}X<{\centering}cX<{\centering}X<{\centering}X<{\centering}}
		\hline
		\multicolumn{1}{c}{Category}& Small 1& \multicolumn{1}{c}{Large } & \multicolumn{1}{c}{DS$^+$} & \multicolumn{1}{c}{Category}& \multicolumn{1}{c}{Small 1}&
		\multicolumn{1}{c}{Large}&\multicolumn{1}{c}{DS$^+$}  \\
		\hline
		Games
		& $84.34\%$          &$96.22\%$&        $96.13\%|88.88\%$       &  Clothing                         & $85.34\%$          &$96.89\%$ & $94.28\%|84.61\%$   \\
		\hline
		Kindle
		& $89.05\%$          &$95.65\%$&        $95.83\%|75.88\%$       &  Beauty                            & $85.37\%$          &$97.20\%$& $94.33\%|86.99\%$    \\
		\hline
		Baby
		& $88.63\%$          &$96.41\%$&        $95.93\%|88.99\%$       &  Video                            & $85.37\%$              & $92.54\%$ & $94.32\%|87.28\%$ \\
		\hline
		Movies
		& $85.37\%$          &$93.42\%$&        $94.23\%|87.68\%$       &   Lawn                           & $85.36\%$              & $89.36\%$ & $94.32\%|94.47\%$   \\
		\hline
		Electronics
		& $85.24\%$          &$95.41\%$&        $94.67\%|88.44\%$       &  Home                         & $85.39\%$              & $96.28\%$ & $94.39\%|88.10\%$   \\
		\hline
		Office
		& $85.23\%$          &$95.45\%$&        $94.68\%|92.12\%$       &  Toys                         & $85.41\%$              & $96.74\%$ & $94.40\%|87.80\%$   \\
		\hline
		CDs
		& $84.68\%$          &$95.87\%$&        $94.86\%|91.99\%$       &  Grocery                            & $85.43\%$              & $96.73\%$ & $94.42\%|89.80\%$   \\
		\hline
		Books
		& $85.26\%$          &$93.66\%$&        $94.20\%|81.88\%$       &  Automotive                            & $85.42\%$              & $94.69\%$ & $94.42\%|90.34\%$   \\
		\hline
		Sports
		& $85.26\%$          &$95.06\%$&        $94.21\%|89.00\%$       &  Tools                         & $85.41\%$              & $94.49\%$ & $94.43\%|90.58\%$   \\
		\hline
		Health
		& $85.24\%$          &$95.04\%$&        $94.23\%|89.49\%$       &  Pet Supplies                          & $85.40\%$          &$94.03\%$ & $94.42\%|90.84\%$   \\
		\hline
		Overall & \multicolumn{3}{c}{Small 1: $85.40\%$, Large: $94.43\%$, DS$^+$: $94.42\%$} &Query & \multicolumn{3}{c}{Small 1: $0\%$, Large: $100\%$, DS$^+$: $84.97\%$}
		\\
		\hline
		\multicolumn{1}{c}{Category}& Small 2& \multicolumn{1}{c}{Large } & \multicolumn{1}{c}{DS$^+$} & \multicolumn{1}{c}{Category}& \multicolumn{1}{c}{Small 2}&
		\multicolumn{1}{c}{Large}&\multicolumn{1}{c}{DS$^+$}  \\
		\hline
		Games
		& $85.29\%$          &$96.22\%$&        $96.13\%|84.01\%$       &  Clothing                         & $86.13\%$          &$96.89\%$ & $94.31\%|74.78\%$   \\
		\hline
		Kindle
		& $89.74\%$          &$95.65\%$&        $95.85\%|71.73\%$       &  Beauty                            & $86.17\%$          &$97.20\%$& $94.36\%|81.93\%$    \\
		\hline
		Baby
		& $89.33\%$          &$96.41\%$&        $95.95\%|82.56\%$       & Video                            & $86.18\%$              & $92.54\%$ & $94.35\%|78.15\%$ \\
		\hline
		Movies 
		& $86.27\%$          &$93.42\%$&        $94.25\%|80.82\%$       &   Lawn                           & $86.17\%$              & $89.36\%$ & $94.35\%|90.64\%$   \\
		\hline
		Electronics
		& $86.28\%$          &$95.41\%$&        $94.69\%|83.57\%$       &  Home                       & $86.21\%$              & $96.28\%$ & $94.41\%|81.93\%$   \\
		\hline
		Office 
		& $86.26\%$          &$95.45\%$&        $94.69\%|89.14\%$       &  Toys                       & $86.22\%$              & $96.74\%$ & $94.43\%|81.05\%$   \\
		\hline
		CDs 
		& $85.70\%$          &$95.87\%$&        $94.88\%|87.78\%$       &  Grocery                            & $86.24\%$              & $96.73\%$ & $94.45\%|84.16\%$   \\
		\hline
		Books
		& $86.05\%$          &$93.66\%$&        $94.23\%|77.59\%$       &  Automotive                            & $86.24\%$              & $94.69\%$ & $94.45\%|87.44\%$   \\
		\hline
		Sports
		& $86.05\%$          &$95.06\%$&        $94.24\%|82.66\%$       &  Tools                         & $86.23\%$              & $94.49\%$ & $94.45\%|86.58\%$   \\
		\hline
		Health
		& $86.03\%$          &$95.04\%$&        $94.26\%|85.12\%$       &  Pet Supplies                          & $86.21\%$          &$94.03\%$ & $94.45\%|86.26\%$   \\
		\hline
		Overall& \multicolumn{3}{c}{Smal 2: $86.21\%$, Large: $94.43\%$, DS$^+$: $94.44\%$} &Query & \multicolumn{3}{c}{Small 2: $0\%$, Large: $100\%$, DS$^+$: $80.00\%$}\\
		\hline
		\multicolumn{1}{c}{Category}& Small 3& \multicolumn{1}{c}{Large } & \multicolumn{1}{c}{DS$^+$} & \multicolumn{1}{c}{Category}& \multicolumn{1}{c}{Small 3}&
		\multicolumn{1}{c}{Large}&\multicolumn{1}{c}{DS$^+$}  \\
		\hline
		Games
		& $90.39\%$          &$96.22\%$&        $96.15\%|36.25\%$       &  Clothing                         & $95.63\%$          &$96.89\%$ & $97.45\%|31.10\%$   \\
		\hline
		Kindle 
		& $95.89\%$          &$95.65\%$&        $97.38\%|20.67\%$       &  Beauty                            & $92.90\%$          &$97.20\%$& $97.24\%|30.39\%$    \\
		\hline
		Baby
		& $92.81\%$          &$96.41\%$&        $96.26\%|32.90\%$       &   Video                            & $92.67\%$              & $92.54\%$ & $96.27\%|25.32\%$ \\
		\hline
		Movies 
		& $90.57\%$          &$93.42\%$&        $94.86\%|31.76\%$       &  Lawn                           & $84.26\%$              & $89.36\%$ & $90.63\%|48.93\%$   \\
		\hline
		Electronics
		& $91.76\%$          &$95.41\%$&        $96.11\%|39.56\%$       &  Home                        & $93.12\%$              & $96.28\%$ & $96.73\%|33.89\%$   \\
		\hline
		Office 
		& $90.72\%$          &$95.45\%$&        $95.10\%|44.31\%$       &  Toys                & $92.22\%$              & $96.74\%$ & $96.38\%|31.34\%$   \\
		\hline
		CDs 
		& $88.57\%$          &$95.87\%$&        $95.50\%|36.92\%$       &  Grocery                            & $92.50\%$              & $96.73\%$ & $96.66\%|32.28\%$   \\
		\hline
		Books
		& $91.98\%$          &$93.66\%$&        $95.37\%|27.91\%$       &  Automotive                            & $91.30\%$              & $94.69\%$ & $94.69\%|33.33\%$   \\
		\hline
		Sports
		& $93.11\%$          &$95.06\%$&        $96.02\%|34.83\%$       &  Tools                         & $91.62\%$              & $94.49\%$ & $95.38\%|37.87\%$   \\
		\hline
		Health
		& $91.73\%$          &$95.04\%$&        $95.71\%|34.35\%$       &  Pet Supplies                          & $91.02\%$          &$94.03\%$ & $95.02\%|38.96\%$   \\
		\hline
		Overall & \multicolumn{3}{c}{Small 3: $91.79\%$, Large: $94.43\%$, DS$^+$: $\bm{95.64\%}$} &Query & \multicolumn{3}{c}{Small 3: $0\%$, Large: $100\%$, DS$^+$: $\bm{31.18\%}$}\\
		\hline
	\end{tabularx}

\end{table*}

\section{Experiments}
In our experiments, we aim to 
(1) validate that DS$^+$ can improve the performance of large models, while reducing the cost across various modalities and tasks, 
(2) validate the effectiveness of PP, PT and 2CD, respectively,
(3) validate that the collaborative-based paradigm can achieve better results compared to fine-tuning,
(4) analyze the important hyperparameter, shunt threshold,
(5) compare confidence with other ways for data shunt.
In our experiments, large models and small models are relative. For example, model A is small compared to model B, but large compared to C. Moreover, the overall performance of large models is always better than that of small models. To save on experimental costs, we utilize the paid large model, ChatGPT, in two experiment, while using relatively smaller free pretrained models as substitutes in the remaining experiments. Therefore, we use the query proportion of large models as an evaluation metric instead of cost. The code for the proposed method are provided for research purposes https://github.com/Anfeather/Data-Shunt.

\subsection{Data Shunt$^+$ for Language Modality}
ChatGPT is one of the most influential large language models today. Running ChatGPT requires at least 350GB GPU memory with specialized infrastructure \cite{zheng2022alpa}, which is far beyond affordable for most product teams. 
Consequently, product teams must invoke the ChatGPT interface to accomplish their desired functionalities. Unfortunately, the exorbitant costs associated with the interface substantially diminish the revenue generated by product teams.

In this section, we show that the proposed method can significantly reduce the cost of calling large models while achieving better overall performance. We first conduct sentiment analysis on Amazon Product Data \cite{he2016ups,mcauley2015image}, where there are 20 categories of product comments along with corresponding positive or negative sentiment labels. In addition, we maintain a balance between positive and negative samples, and divide the dataset into training set, validation set, and testing set, with 2,504,958, 277,508, and 309,186 samples respectively. 

Related results are presented in Table \ref{DS_on_AS_}. For DS, the right value of ``$|$" is the sample proportion processed by ChatGPT, while the left value of ``$|$" is the accuracy of DS.
In this experiment, we use TextCNN \cite{kim2014convolutional}, LSTM \cite{wang2016attention} and fine-tuned BERT \cite{devlin2018BERT} as the small model respectively, while ChatGPT serves as the large model. 
It can be seen that the overall accuracy of TextCNN, LSTM, fine-tuned BERT and ChatGPT is $85.40\%$, $86.21\%$, $91.79\%$ and $94.43\%$. 
The pretrained large model, ChatGPT, significantly outperforms specific small models, TextCNN, LSTM, and fine-tuned BERT. Its extensive pretraining equips it with a wealth of knowledge, enabling it to handle a broader range of scenarios.
On the other hand, TextCNN, LSTM and fine-tuned BERT see much data that belongs to the 20 classes, thus, small models may fit some data better. 
For instance, in the Kindle, certain e-books share similar highlights, such as well-developed characterization and compelling storylines.
Thus, during inference, if there is a comment that is similar to one of the training dataset, small models will give a more accurate prediction.
Based on this idea, we have observed that the combination of small and large models can yield competitive or even superior performance. 
As shown in DS$^+$ of Table \ref{DS_on_AS_}, we present results of DS$^+$ combined with three small models. To achieve competitive results to ChatGPT, DS$^+$ with TextCNN needs to send $84.97\%$ of the data to ChatGPT, while DS$^+$ with LSTM needs to send $80.00\%$, and DS$^+$ with fine-tuned BERT needs to send $31.18\%$.
It is evident that as the performance of the small model improves, the required query proportion to attain competitive performance decreases.
Furthermore, DS$^+$ with fine-tuned BERT exhibits remarkable results, benefiting from its exposure to lots of diverse data as well as task-specific data.
Fine-tuned BERT even slightly performs better on two fine-grained classes, Kindle and Video, where the query proportion of DS$^+$ is lower than $26\%$. It demonstrates that DS$^+$ with small foundation models can achieve much better performance. Compared to ChatGPT, DS$^+$ displays a $1.21\%$ increase in accuracy while only  $31.18\%$ of the data is calculated by the large model. 
In terms of actual expenses, employing the large model for computations on 309,186 test samples resulted in a total cost of $\$1,201.87$. Alternatively, using DS reduced the cost to approximately $\$374.74$.


\begin{table}[h]
	\centering
	\caption{Experiments on ChaosNLI.}
	\label{ChaosNLI}
	\begin{tabularx}{\linewidth}{cX<{\centering}X<{\centering}X<{\centering}}
		\hline
		\multicolumn{1}{c}{Model}&Query& \multicolumn{1}{c}{Acc} & \multicolumn{1}{c}{DS$^+$}\\
		\hline
		BERT-base & $75.36\%$          &$55.91\%$&        ${63.79\%}$        \\ \hline
		BERT-large
		& $66.98\%$          &$56.91\%$&        ${63.29\%}$        \\
		\hline
		BART
		& $59.85\%$          &$59.22\%$&        ${65.48\%}$       \\
		\hline
		XLNET-base		& $68.48\%$          &$58.91\%$&        ${64.67\%}$         \\ \hline
		XLNET-large
		& $\bm{58.54\%}$          &$\bm{61.85\%}$&        $\bm{67.29\%}$         \\
		\hline
		ALBERT-xxlarge
		& ${62.48\%}$          &$58.97\%$&        $64.73\%$        \\
		\hline
		DISTILBERT
		& ${85.99\%}$          &$51.03\%$&        $64.23\%$        \\
		\hline
		\multicolumn{1}{c}{ChatGPT}
		&\multicolumn{3}{c}{$63.73\%$}        \\
		\hline
	\end{tabularx}
\end{table}

We further conduct experiments on the ChaosNLI \cite{ynie2020chaosnli} dataset that with inherent ambiguities, and the results are presented in the Table \ref{ChaosNLI}. The large model is ChatGPT, the small model is 
BERT-base, BERT-large \cite{devlin2018BERT}, BART \cite{lewis2019bart}, XLNET-base, XLNET-large \cite{yang2019xlnet}, ALBERT-xxlarge \cite{lan2019albert}, DISTILBERT \cite{sanh2019distilbert}, respectively. To demonstrate the powerful adaptability of the proposed paradigm, no modifications are made to the small model in this experiment.
The related results also validate the effectiveness of DS$^+$ in improving the performance of large models while reducing costs.
Nevertheless, Table \ref{ChaosNLI} reveals two noteworthy observations. 
1. XLNET-base surpasses BERT-large and exhibit similar performance to ALBERT-xxlarge and BART, but the query proportion of XLNET-base is much higher.
This suggests that the performance of the small model is not solely responsible for determining the proportion of data delivered to the large model. 
2. DISTILBERT exhibits significantly poorer performance compared to other small models, yet it remains competitive in the paradigm of DS$^+$.
This indicates that DISTILBERT achieves higher accuracy on high-confidence samples, enabling DS$^+$ to achieve good results even when the small model’s performance is poor.
This phenomenon is likely caused by overfitting, as DISTILBERT exclusively recognizes a specific distribution, which contributes to its overall inferior. However, this characteristic also enables DISTILBERT to accurately determine whether a sample should be processed by the large model. Exploring ways to leverage the phenomenon of overfitting could be a potential research direction in this field. 
Additionally, Table \ref{ChaosNLI} shows that DS$^+$ continues to work well even with small models achieving an accuracy of less around $60\%$.

\subsection{Data Shunt$^+$ for Vision Modality}
In this section, to better present the application of DS$^+$, we conduct experiments for long-tailed image classification. Long-tailed image classification is a common challenge in practical computer vision applications \cite{zhou2018deep}. We follow \cite{li2022trustworthy} to separate CIFAR-100 into the head, medium and tail regions based on different numbers of samples. 

\begin{table}[h]
	\centering
		\caption{DS$^+$ for image classification on CIFAR-100-LT.}
	\label{image_classification}
\begin{tabularx}{\linewidth}{cX<{\centering}X<{\centering}X<{\centering}}
		\hline
		\multicolumn{1}{c}{}& Small & \multicolumn{1}{c}{Large } & \multicolumn{1}{c}{DS$^+$}\\
		\hline
		Head
		& $70.25\%$          &$60.00\%$&        $\bm{71.99\%}$       \\
		\hline
		Med
		& $46.61\%$          &$57.28\%$&        $\bm{59.91\%}$         \\
		\hline
		Tail
		& $29.28\%$          &$57.19\%$&        $\bm{57.61\%}$        \\
		\hline
		Overall Accuracy
		& $48.84\%$          &$58.18\%$&        $\bm{63.25\%}$        \\
		\hline
		Query Proportion 
		& $\bm{0\%}$          &$100\%$&        $66.10\%$        \\
		\hline
	\end{tabularx}

\end{table}

Due to budget constraints and the limited availability of vision pretrained models, in this experiment, we regard ResNet-32 \cite{he2016deep} as the small model, CLIP as the large model. The results on CIFAR-100-LT are show in Table \ref{image_classification}. 
Different from the experiment of language modality in the previous section,  the performance of the small model is greatly affected by the number of training samples. The small model achieving $70.25\%$, $46.61\%$ and $29.28\%$ accuracy in the head, medium, and tail regions, respectively.
The accuracy of the small model for the head data even significantly surpasses that of the large models, $60\%$, which further highlights that small models and large models can have their respective advantages. 
In contrast, small models yield similar results across most classes in the language modality. These outcomes might stem from a higher degree of similarity between comments, in contrast to images of different classes.
Compared to established baselines that solely rely on either the small model or the large model, our approach yields significant performance improvements in all cases, especially, the overall accuracy has been improved by $5.07\%$ compared to the large model, while the number of querying reduces to $66.10\%$ of the original.

\subsection{Data Shunt$^+$ for Multimodality}
In this section, we validate the effectiveness of the proposed DS$^+$ on the image caption task \cite{stefanini2022show}, which is a generative task, different from previous experiments.

We follow \cite{xu2015show} to design the small model, where the encoder is a ResNet-101, and the decoder is an LSTM. To get the confidence of the small model for the input sample, we calculate the mean of probabilities that predict next words. 

The confidence of the small model for an input image is:
\begin{equation}
\begin{aligned}
\mathbf{i}_t & =\sigma\left(W_i E \mathbf{y}_{t-1}+U_i \mathbf{h}_{t-1}+Z_i \hat{\mathbf{z}}_t+\mathbf{b}_i\right), \\
\mathbf{f}_t & =\sigma\left(W_f E \mathbf{y}_{t-1}+U_f \mathbf{h}_{t-1}+Z_f \hat{\mathbf{z}}_t+\mathbf{b}_f\right), \\
\mathbf{c}_t & =\mathbf{f}_t \mathbf{c}_{t-1}+\mathbf{i}_t \tanh \left(W_c E \mathbf{y}_{t-1}+U_c \mathbf{h}_{t-1}+Z_c \hat{\mathbf{z}}_t+\mathbf{b}_c\right), \\
\mathbf{o}_t & =\sigma\left(W_o E \mathbf{y}_{t-1}+U_o \mathbf{h}_{t-1}+Z_o \hat{\mathbf{z}}_t+\mathbf{b}_o\right), \\
\mathbf{h}_t & =\mathbf{o}_t \odot \tanh \left(\mathbf{c}_t\right),\\
C_t &= \frac{e^{z_i}}{\sum{e_{z_d}}}, \quad z_i \in F_{MLP}(\mathbf{h}_t) .
\end{aligned}
\label{confidence_caption}
\end{equation}
where $\mathbf{i}_t$, $\mathbf{f}_t$, $\mathbf{c}_t$, $\mathbf{o}_t$, $\mathbf{h}_t$ are the input, forget, memory, output and hidden state of the LSTM, respectively. $\hat{\mathbf{z}}$ is the context vector, capturing the visual information, as explained below. $E$ is an embedding matrix. $F_{MLP}$ is a full connected network. $\sigma$ and $\odot$ is the logistic sigmoid activation and element-wise multiplication, respectively.

\begin{table}[h]
	\centering
		\caption{DS$^+$ for image caption on Microsoft COCO.}
	\label{image_caption}
\begin{tabularx}{\linewidth}{cX<{\centering}X<{\centering}X<{\centering}}
		\hline
		\multicolumn{1}{c}{}& Small & \multicolumn{1}{c}{Large } & \multicolumn{1}{c}{DS$^+$}\\
		\hline
		BLEU-1 & $72.92$          &$73.27$&        $\bm{74.95}$        \\
		\hline
		BLEU-2 & $55.73$          &${60.04}$&        $\bm{60.43}$        \\
		\hline
		BLEU-3 & $41.20$          &$\bm{46.99}$&        ${46.85}$        \\
		\hline
		BLEU-4 & $30.28$          &$\bm{36.11}$&        ${35.82}$        \\
		\hline
		Mean
		& ${50.03}$          &${54.10}$&        $\bm{54.52}$        \\
		\hline
		Query Proportion 
		& $\bm{0\%}$          &$100\%$&        $65.36\%$        \\
		\hline
	\end{tabularx}
\end{table}
The image caption experiments are conducted on Microsoft COCO \cite{lin2014microsoft}, which comprises 82,783 images with captions. We follow \cite{xu2015show} to devide the training, validation, and testing set. As for the large model, we employ BLIP-2 (1.1B) \cite{li2023blip}. In this experiment, when the confidence computed by Equation \ref{confidence_caption} is lager than 0.55, the input data will be processed by the small model, otherwise it is processed by the large model.
Related results about BLEU \cite{papineni2002bleu} are reported in Table \ref{image_caption}. 
It can be observed that the small model is inferior to the large model in terms of every metric. Additionally, the small model has much poorer ability to generate fluent sentences compared to the large model, as the difference between the small and large models becomes more significant with the n-gram (BLEU-n) increasing. 
As for DS$^+$, it can leverage the strengths of both the small model and the large model, resulting in improved performance on BLEU-1 and BLEU-2. However, when it comes to BLEU-3 and BLEU-4, which involve longer word combinations, DS$^+$ falls short of surpassing the performance of the large model.
BLEU-1 is closely related to the previous classification task, in which it predicts the likelihood of specific words appearing based on images. It can be found that the proposed DS$^+$ yields better performance in prediction tasks. This might be attributed to DS$^+$ directing data flow based on the confidence of the predictions. 
Nonetheless, this experiment still validates the effectiveness of the proposed DS$^+$ for image caption task, as DS$^+$ successfully improve in the average BLEU score, while solely $65.36\%$ of the data is computed by the large model.

\subsection{Ablation for PP and 2CD}
In this section, we conduct ablation experiments to show the effectiveness of PP and 2CD on CIFAR-100-LT. Related resulrs are shown in Table \ref{ablation_a}.

\begin{table}[h]
	\centering
	\caption{Ablation experiments of DS$^+$ on CIFAR-100-LT.}
	\label{ablation_a}
	\begin{tabularx}{\linewidth}{cX<{\centering}X<{\centering}X<{\centering}}
		\hline
		\multicolumn{1}{c}{}& DS$^+$ & \multicolumn{1}{c}{DS$^+$-2CD } & \multicolumn{1}{c}{DS$^+$-PP-2CD}\\
		\hline
		Head
		& $\bm{71.99\%}$          &$71.21\%$&        $71.54\%$       \\
		\hline
		Med
		& $\bm{59.91\%}$           &$58.69\%$&        $59.76\%$         \\
		\hline
		Tail
		& $\bm{57.61\%}$          &$56.17\%$&        $53.31\%$        \\
		\hline
		Overall Accuracy
		& $\bm{63.25\%}$         &$62.11\%$&        $61.63\%$        \\
		\hline
		Query Proportion 
		& $\bm{66.10\%}$          &$67.48\%$&        $67.48\%$        \\
		\hline
	\end{tabularx}
	
\end{table}

It can be seen that both PP and 2CD have a positive impact on the proposed method. Besides, according to the Query Proportion of DS$^+$, $66.10\%$, and DS$^+$-2CD, $67.48\%$, 2CD can further reduce the number of times calling the large model, as small models have learned more data distributions. 
Moreover, from the comparison between DS$^+$-2CD and DS$^+$-2CD-PP, we find that PP primarily works on the tail data, as the accuracy improved by $2.86\%$.
It is in line with our previous idea in section \textit{Small Models for Large Models, Prompt Pruning}, with the prior knowledge of the small model, PP can reduce the candidate classes and improve the accuracy of the tail data.

\subsection{Improving Large Models with PT}

\begin{figure}[h]
	\centering
	\subfloat[without PT]{\includegraphics[width=1.7in]{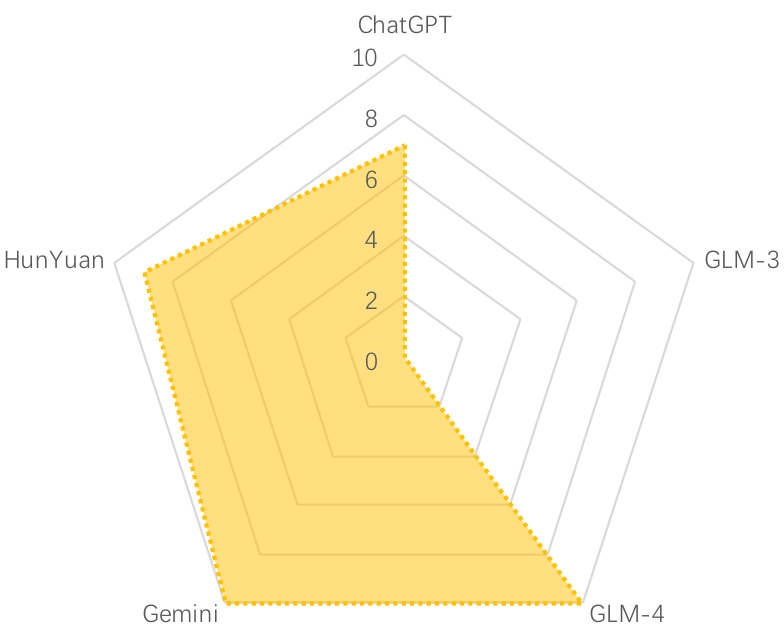}%
		\label{PT_result_1}}
	\hfil
	\subfloat[with PT]{\includegraphics[width=1.7in]{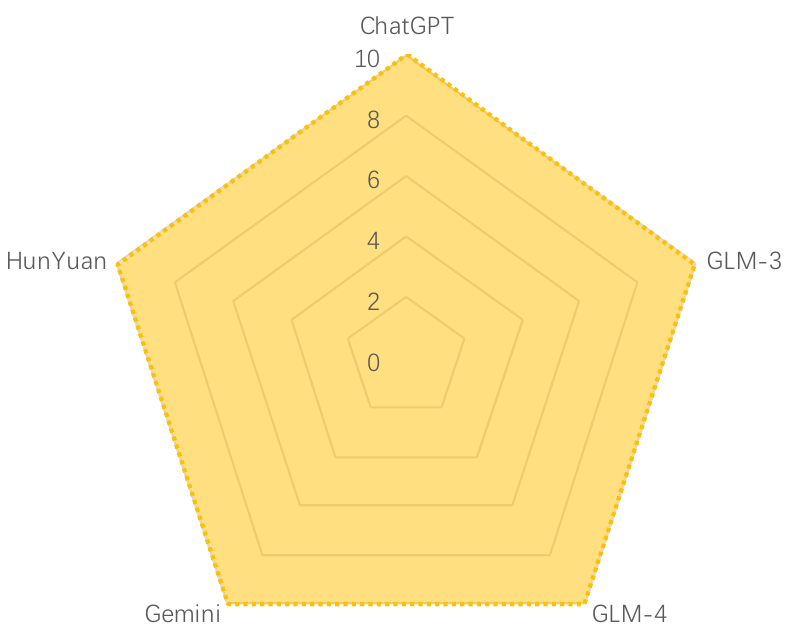}%
		\label{PT_result_2}}
	\caption{Results of paid large models with or without PT.}
	\label{PT_result}
\end{figure}

To validate the effectiveness of PT, we chose a complex application scenario, dispute liability determination, where a narrator describes their dispute with another party, and a large language model is employed to determine the greater responsibility of each party involved. For this experiments, the large language models are ChatGPT, GLM-3, GLM-4, HunYuan and Gemini. 
We conducted a case study in which each model performs ten analyses of the same case. The corresponding results are reported in Fig. \ref{PT_result}. As shown in Fig. \ref{PT_result_1}, only GLM-4 and Gemini achieve perfect scores, while GLM-3 was incorrect in all ten attempts. Such results arise from the scenario of dispute liability determination, where narrators tend to focus more on the aspects that are favorable to themselves, leading the large model to overlook the core elements of the event. PT divides dispute liability determination into subtasks: redundancy removal and dispute adjudication. The redundancy removal subtask employs a small model (ChatGLM3-6B) to abstract the narrator's input and eliminate redundant information, while the dispute adjudication subtask is handled by large models. By using a small model complete simple tasks, PT allows the large model to focus better on the key elements of the task, thereby improving the performance. Results with PT are illustrated in Fig. \ref{PT_result_2}, where each large model can achieve a perfect score.

We present the case study in Table \ref{case}. Related results show that without PT, large models tend to focus on interference information introduced by the narrator, leading to incorrect judgments. By employing small models to denoise the narrator's input, PT can effectively redirect the attention of the large models to the key elements of the event.

 In addition, the reduction in the number of input tokens due to the elimination of redundant information will significantly reduce the querying costs of large models. In this case, the number of input tokens by the narrator is 216, and it is reduced to 53 after being processed by the small model.  

\begin{table*}[!htbp] 
	\centering
	\caption{Case stuty of dispute liability determination.}
	\label{case}
	\begin{tabular}{l} 
		\toprule 
		\textbf{\textit{Narrator: }}\\
		Early that day, Xiao Ming from next door came to my house carrying his household garbage and placed the trash bag at\\ my doorstep. Just as I was coming out of my house, I saw the garbage bag and became very angry, questioning why he had\\ thrown the garbage at my doorstep. However, Xiao Ming insisted that he had placed the garbage bag in the corridor in front\\ of his own house, emphasizing that it was none of my business. The spot where he placed the garbage bag was less than ten\\ centimeters away from my doorstep, how could that be considered as being in front of his own house? Additionally, garbage\\ should be directly disposed of in the downstairs trash bin instead of being piled up upstairs. This not only creates a dirty and\\ unsightly environment but also becomes a breeding ground for insects, especially in the summer. Furthermore, by leaving the\\ garbage in the corridor, it poses a significant safety hazard in the event of a fire or other emergencies. I argued with him for a\\ while, and in the end, I couldn’t contain my anger and ended up punching him.\\
		\hline
		\noindent\textit{\textbf{Summarization (by ChatGLM3-6B):}} \\
		Xiao Ming, a neighbor, placed his garbage bag at the narrator's doorstep instead of in the corridor in front of his own house.\\ The narrator was angry and argued with Xiao Ming, but Xiao Ming refused to take responsibility. The narrator ended up\\ punching him.
		\\
		\hline
		\textbf{\textit{Prompt (for paid large models):}} \\
		Based on the dispute statement below, determine which party is more at fault in the incident (you must explicitly name the\\ party more at fault, the narrator or the other party. Only indicate the party that made the more severe mistake, without\\ providing any other analysis content): \\
		\hline
		\textbf{\textit{Response from GLM-3 (without PT):}} \\
		In this incident, the other party (Xiao Ming) is more at fault. \textbf{\textit{He incorrectly placed the garbage bag in a location that was less}}\\ \textbf{\textit{than ten centimeters away from the narrator's doorstep, which can be seen as disrespectful and inconsiderate. Additionally,}}\\ \textbf{\textit{leaving garbage in the corridor can create an unsightly environment and poses safety hazards.}} The narrator's response of\\ punching Xiao Ming is also not justified, but the initial mistake made by Xiao Ming makes him more at fault in this situation.\\
		\hline
		\textbf{\textit{Response from GLM-3 (with PT):}} \\
		In this scenario, the party more at fault is the narrator. The narrator's response to the situation involving the garbage bag was\\ disproportionate and violent. While it is understandable that the narrator might be frustrated or angered by Xiao Ming's actions,\\ resorting to physical violence is not justified and is against the principles of conflict resolution and maintaining community\\ harmony. The narrator's decision to punch Xiao Ming escalated the situation severely and is the more severe mistake in this\\ incident.\\
		\bottomrule 
	\end{tabular}
\end{table*}

\subsection{Collaboration VS Fine-tuning}
In this section, we compare the proposed collaborative-based paradigm with the most widely used fine-tuning method, LORA \cite{hu2021lora}. The large model is LLama3-8b-instruct \cite{meta2024introducing}, the small model is XLNET-large, experiments are conducted on ChaosNLI.

\begin{figure}[h]
	\centering
	\includegraphics[scale=0.65]{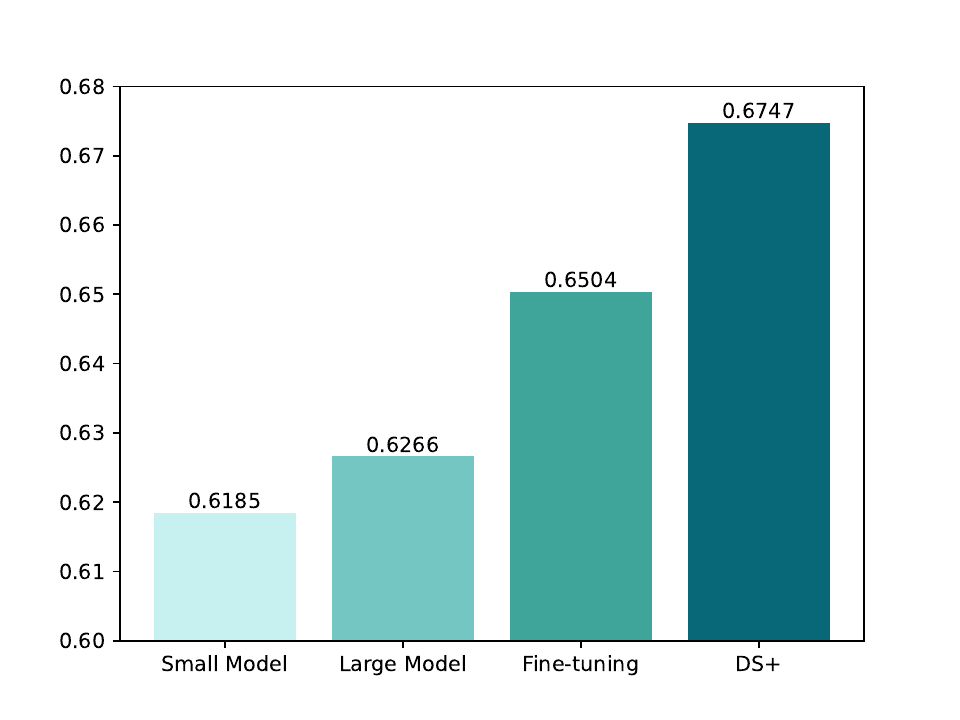}
	\caption{Comparison between fine-tuning and collaborative-based paradigm. The large model is LLama3-8b-instruct, the small model is XLNET-large.}
	\label{compare_with_FT}
\end{figure}
The large model is trained on 160,000 samples for fine-tuning (we have also used other amounts of data to fine-tune the large model, but all the results are inferior to that reported in Fig. \ref{compare_with_FT}).
As shown in Fig. \ref{compare_with_FT}, after fine-tuning, the performance of the large model has improved by $2.38\%$, indicating that fine-tuning can effectively inject specific knowledge of related tasks. On the other hand, while the accuracy of the small model is only $61.85\%$, and the accuracy of the large model without fine-tuning is $62.66\%$, the $DS^+$ achieves $67.47\%$ by leveraging the cooperation between the small and large models to complement each other's deficiencies, significantly surpassing the accuracy of the fine-tuned large model.

\subsection{Hyperparameter Analysis}

This section primarily focuses on the important hyperparameter, shunt threshold, $\delta$, which governs the data flow. Specifically, when the confidence of a sample is larger than $\delta$, this sample will solely be processed by small models, otherwise, this sample will be processed by large models.
To better present the influence of $\delta$, we conduct related experiments on sentiment analysis based on the prior section.
\begin{figure}[h]
	\centering
	\includegraphics[scale=0.55]{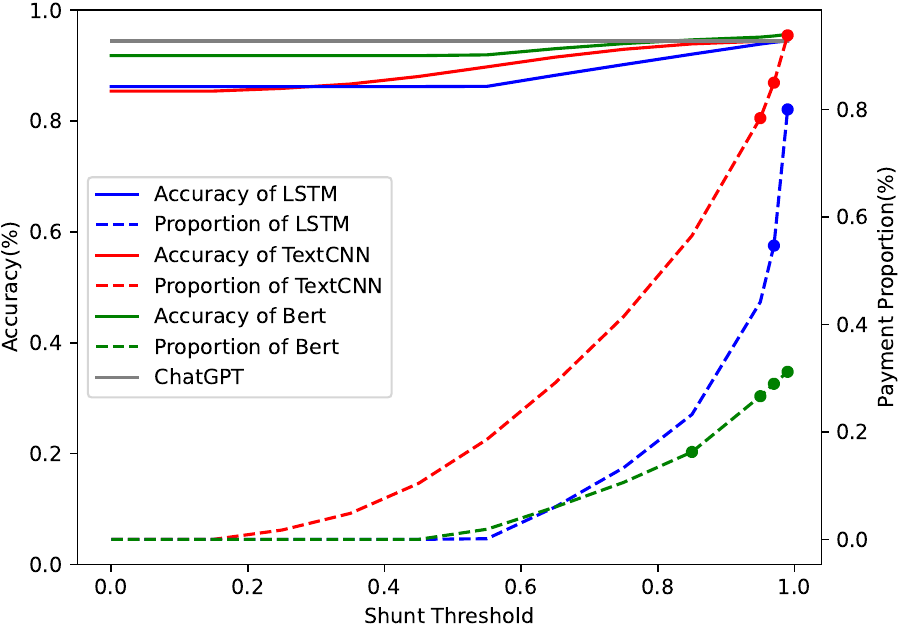}
	\caption{Hyperparameter analysis for shunt threshold on sentiment analysis. The solid line represents the accuracy of DS$^+$ with different small models. The dotted line represents the proportion of samples computed by the large model. The bold dot represents DS$^+$ achieves better performance than that of the large model.}
	\label{hyperparameter_data_shunt}
\end{figure}

As illustrated in Fig. \ref{hyperparameter_data_shunt}, DS$^+$ with three different small models all can surpass the large model. 
When DS$^+$ surpasses the large model, the requirement for the hyperparameter $\delta$ becomes lower for better-performing small models (i.e., $\delta$ can have a wider range). 
For example, DS$^+$ with TextCNN or LSTM requires $\delta>0.97$, while DS$^+$ with fine-tuned BERT requires $\delta>0.85$. Besides, when $\delta>0.97$,  the overall performance of DS$^+$ shows only a slight improvement, while the proportion of samples processed by the large model increases significantly. 
For instance, when $\delta = 0.97$ and $0.99$, the accuracy of DS$^+$ with LSTM is $94.20\%$ and $94.45\%$, respectively. As the parameter $\delta$ increases, the change in accuracy is minimal, while the query proporyion has significantly increased from $54.66\%$ to $80.00\%$. Although such phenomenon is not as prominent for DS$^+$ with TextCNN (i.e., the vertical distance between the two points on the right side of the red dashed line) and DS$^+$ with BERT (i.e., the vertical distance between the two points on the right side of the green dashed line), it is important to select a $\delta$ between $[0.97,0.99]$ on the validation dataset.


\subsection{Why Shunting with Confidence}
In addition to using confidence levels, data can be shunted with other methods. One approach is to introduce a new model that predicts the distribution, or alternatively, predict the data flow based on the small model’s performance on the training data. 
This section will compare different ways of shunting in the context of image classification. For shunting with distributions, we train a distribution model to classify inputs into categories such as head, med, or tail, which represents a coarser-grained classification. For direct prediction, we train a model that predicts the accuracy of the small model during training.

\begin{figure}[h]
	\centering
	\includegraphics[scale=0.55]{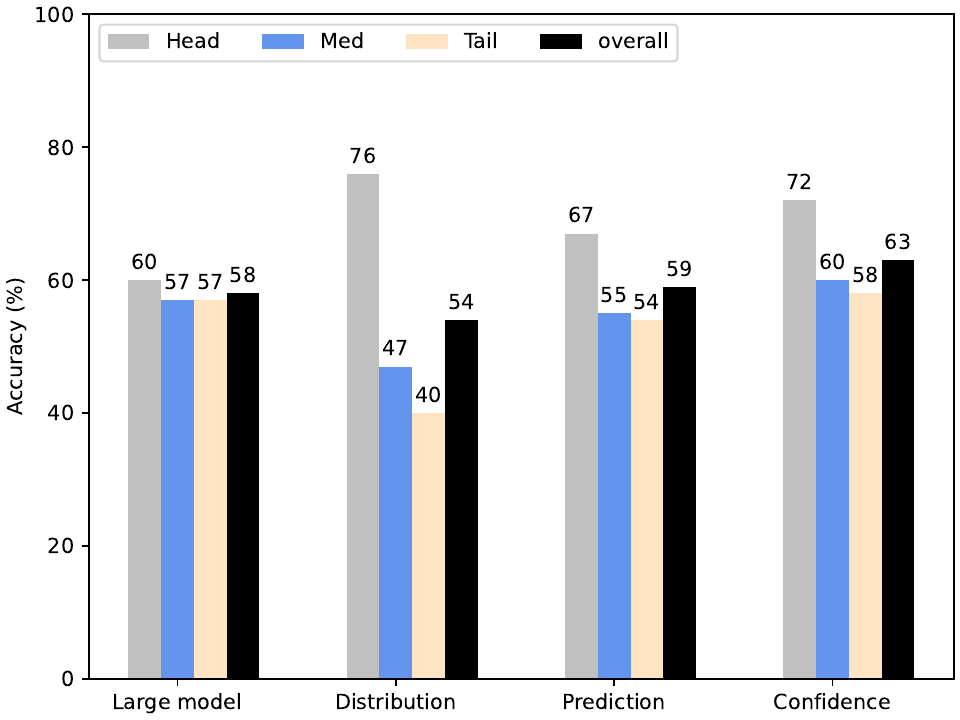}
	\caption{Different shunting ways of DS$^+$.}
	\label{why_confidence}
\end{figure}

As demonstrated in Fig. \ref{why_confidence}, shunting with distributions significantly degrades the overall performance compared to that of the large model. Importantly, this approach shows a particularly noticeable improvement for the head data. This is because the distribution model assigns data that aligns with the head distribution of the training set to the small model, while allocating the head data that deviates from it (or hard head samples) to the large model.
As for shunting with prediction, it only has a slight improvement compared to the large model, because of the inaccuracy of the prediction model that predicts whether the small model can give the correct answer.
In contrast, shunting with confidence does not introduce an extra model and the performance is much more better. During inference, when an input data is similar to the training data, the small model will exhibit high confidence and is more likely to handle it correctly. Conversely, if the small model encounters an unfamiliar sample, its confidence level will be low, making it more suitable for the large model to handle.

\section{Conclusion}
With the advancements in PLMs, an increasing number of related applications are gradually becoming integrated into people’s daily lives.
The enormous computational resources required by PLMs have deterred the majority of product teams. 
Utilizing PLMs through interface will incur significant costs. Therefore, we propose Data Shunt$^+$ (DS$^+$), a general paradigm for collaboration of small and large models, which not only substantially reduces the cost associated with querying large models but also effectively improves large models' performance. Specifically, DS$^+$ comprises two primary modules: Small Model for Large Model (S4L) and Large Model for Small Model (L4S). In S4L, we introduce Prompt Pruning (PP) and Prompt Transferring (PT), which refine the prediction space and the input space of large models, respectively. Regarding L4S, we introduce 2-Stage Confidence Distillation (2CD), which integrates the knowledge of large models into small models and prevents the small models from forgetting the knowledge acquired during training. The remarkable performance across diverse modalities and tasks demonstrates the superiority of the proposed DS$^+$. 


%

\ifCLASSOPTIONcompsoc
  \section*{Acknowledgments}
\else
  \section*{Acknowledgment}
\fi

This work is supported by the National Key Research and Development Program of China (grant no.2021YFB3301504), National Key R$\&$D Program of China under Grant 2022ZD0160101, NSFC (No. 62272411), the Zhejiang NSF (LR21F020004), and Ant Group, My Bank.

\ifCLASSOPTIONcaptionsoff
  \newpage
\fi



%
	\bibliographystyle{IEEEtran}
\bibliography{IEEEabrv,IEEE}

\begin{thebibliography}{10}
\providecommand{\url}[1]{#1}
\csname url@samestyle\endcsname
\providecommand{\newblock}{\relax}
\providecommand{\bibinfo}[2]{#2}
\providecommand{\BIBentrySTDinterwordspacing}{\spaceskip=0pt\relax}
\providecommand{\BIBentryALTinterwordstretchfactor}{4}
\providecommand{\BIBentryALTinterwordspacing}{\spaceskip=\fontdimen2\font plus
\BIBentryALTinterwordstretchfactor\fontdimen3\font minus
  \fontdimen4\font\relax}
\providecommand{\BIBforeignlanguage}[2]{{%
\expandafter\ifx\csname l@#1\endcsname\relax
\typeout{** WARNING: IEEEtran.bst: No hyphenation pattern has been}%
\typeout{** loaded for the language `#1'. Using the pattern for}%
\typeout{** the default language instead.}%
\else
\language=\csname l@#1\endcsname
\fi
#2}}
\providecommand{\BIBdecl}{\relax}
\BIBdecl

\bibitem{yu2023hallucidoctor}
Q.~Yu, J.~Li, L.~Wei, L.~Pang, W.~Ye, B.~Qin, S.~Tang, Q.~Tian, and Y.~Zhuang,
  ``Hallucidoctor: Mitigating hallucinatory toxicity in visual instruction
  data,'' \emph{arXiv preprint arXiv:2311.13614}, 2023.

\bibitem{li2023gradient}
J.~Li, M.~Gao, L.~Wei, S.~Tang, W.~Zhang, M.~Li, W.~Ji, Q.~Tian, T.-S. Chua,
  and Y.~Zhuang, ``Gradient-regulated meta-prompt learning for generalizable
  vision-language models,'' 2023.

\bibitem{bommasani2021opportunities}
R.~Bommasani, D.~A. Hudson, E.~Adeli, R.~Altman, S.~Arora, S.~von Arx, M.~S.
  Bernstein, J.~Bohg, A.~Bosselut, E.~Brunskill \emph{et~al.}, ``On the
  opportunities and risks of foundation models,'' \emph{arXiv preprint
  arXiv:2108.07258}, 2021.

\bibitem{brown2020language}
T.~Brown, B.~Mann, N.~Ryder, M.~Subbiah, J.~D. Kaplan, P.~Dhariwal,
  A.~Neelakantan, P.~Shyam, G.~Sastry, A.~Askell \emph{et~al.}, ``Language
  models are few-shot learners,'' \emph{Advances in neural information
  processing systems}, vol.~33, pp. 1877--1901, 2020.

\bibitem{ouyang2022training}
L.~Ouyang, J.~Wu, X.~Jiang, D.~Almeida, C.~Wainwright, P.~Mishkin, C.~Zhang,
  S.~Agarwal, K.~Slama, A.~Ray \emph{et~al.}, ``Training language models to
  follow instructions with human feedback,'' \emph{Advances in Neural
  Information Processing Systems}, vol.~35, pp. 27\,730--27\,744, 2022.

\bibitem{liu2023pre}
P.~Liu, W.~Yuan, J.~Fu, Z.~Jiang, H.~Hayashi, and G.~Neubig, ``Pre-train,
  prompt, and predict: A systematic survey of prompting methods in natural
  language processing,'' \emph{ACM Computing Surveys}, vol.~55, no.~9, pp.
  1--35, 2023.

\bibitem{wei2022emergent}
J.~Wei, Y.~Tay, R.~Bommasani, C.~Raffel, B.~Zoph, S.~Borgeaud, D.~Yogatama,
  M.~Bosma, D.~Zhou, D.~Metzler \emph{et~al.}, ``Emergent abilities of large
  language models,'' \emph{arXiv preprint arXiv:2206.07682}, 2022.

\bibitem{zhu2023visual}
J.~Zhu, S.~Lai, X.~Chen, D.~Wang, and H.~Lu, ``Visual prompt multi-modal
  tracking,'' in \emph{Proceedings of the IEEE/CVF Conference on Computer
  Vision and Pattern Recognition}, 2023, pp. 9516--9526.

\bibitem{alayrac2022flamingo}
J.-B. Alayrac, J.~Donahue, P.~Luc, A.~Miech, I.~Barr, Y.~Hasson, K.~Lenc,
  A.~Mensch, K.~Millican, M.~Reynolds \emph{et~al.}, ``Flamingo: a visual
  language model for few-shot learning,'' \emph{Advances in Neural Information
  Processing Systems}, vol.~35, pp. 23\,716--23\,736, 2022.

\bibitem{li2023blip}
J.~Li, D.~Li, S.~Savarese, and S.~Hoi, ``Blip-2: Bootstrapping language-image
  pre-training with frozen image encoders and large language models,''
  \emph{arXiv preprint arXiv:2301.12597}, 2023.

\bibitem{zhang2024vision}
J.~Zhang, J.~Huang, S.~Jin, and S.~Lu, ``Vision-language models for vision
  tasks: A survey,'' \emph{IEEE Transactions on Pattern Analysis and Machine
  Intelligence}, 2024.

\bibitem{zeng2023x}
Y.~Zeng, X.~Zhang, H.~Li, J.~Wang, J.~Zhang, and W.~Zhou, ``X 2-vlm: All-in-one
  pre-trained model for vision-language tasks,'' \emph{IEEE Transactions on
  Pattern Analysis and Machine Intelligence}, 2023.

\bibitem{xu2023multimodal}
P.~Xu, X.~Zhu, and D.~A. Clifton, ``Multimodal learning with transformers: A
  survey,'' \emph{IEEE Transactions on Pattern Analysis and Machine
  Intelligence}, 2023.

\bibitem{surameery2023use}
N.~M.~S. Surameery and M.~Y. Shakor, ``Use chat gpt to solve programming
  bugs,'' \emph{International Journal of Information Technology \& Computer
  Engineering (IJITC) ISSN: 2455-5290}, vol.~3, no.~01, pp. 17--22, 2023.

\bibitem{biswas2023role}
S.~Biswas, ``Role of chat gpt in education,'' \emph{Available at SSRN 4369981},
  2023.

\bibitem{ouyang2016factors}
W.~Ouyang, X.~Wang, C.~Zhang, and X.~Yang, ``Factors in finetuning deep model
  for object detection with long-tail distribution,'' in \emph{Proceedings of
  the IEEE conference on computer vision and pattern recognition}, 2016, pp.
  864--873.

\bibitem{zhang2017range}
X.~Zhang, Z.~Fang, Y.~Wen, Z.~Li, and Y.~Qiao, ``Range loss for deep face
  recognition with long-tailed training data,'' in \emph{Proceedings of the
  IEEE International Conference on Computer Vision}, 2017, pp. 5409--5418.

\bibitem{oksuz2020imbalance}
K.~Oksuz, B.~C. Cam, S.~Kalkan, and E.~Akbas, ``Imbalance problems in object
  detection: A review,'' \emph{IEEE transactions on pattern analysis and
  machine intelligence}, vol.~43, no.~10, pp. 3388--3415, 2020.

\bibitem{horenko2020scalable}
I.~Horenko, ``On a scalable entropic breaching of the overfitting barrier for
  small data problems in machine learning,'' \emph{Neural Computation},
  vol.~32, no.~8, pp. 1563--1579, 2020.

\bibitem{smith2022using}
S.~Smith, M.~Patwary, B.~Norick, P.~LeGresley, S.~Rajbhandari, J.~Casper,
  Z.~Liu, S.~Prabhumoye, G.~Zerveas, V.~Korthikanti \emph{et~al.}, ``Using
  deepspeed and megatron to train megatron-turing nlg 530b, a large-scale
  generative language model,'' \emph{arXiv preprint arXiv:2201.11990}, 2022.

\bibitem{zhang2022opt}
S.~Zhang, S.~Roller, N.~Goyal, M.~Artetxe, M.~Chen, S.~Chen, C.~Dewan, M.~Diab,
  X.~Li, X.~V. Lin \emph{et~al.}, ``Opt: Open pre-trained transformer language
  models,'' \emph{arXiv preprint arXiv:2205.01068}, 2022.

\bibitem{hoffmann2022training}
J.~Hoffmann, S.~Borgeaud, A.~Mensch, E.~Buchatskaya, T.~Cai, E.~Rutherford,
  D.~d.~L. Casas, L.~A. Hendricks, J.~Welbl, A.~Clark \emph{et~al.}, ``Training
  compute-optimal large language models,'' \emph{arXiv preprint
  arXiv:2203.15556}, 2022.

\bibitem{li2023finetuning}
J.~Li, K.~Pan, Z.~Ge, M.~Gao, H.~Zhang, W.~Ji, W.~Zhang, T.-S. Chua, S.~Tang,
  and Y.~Zhuang, ``Fine-tuning multimodal llms to follow zero-shot
  demonstrative instructions,'' \emph{arXiv preprint arXiv:2308.04152}, 2023.

\bibitem{zheng2022alpa}
L.~Zheng, Z.~Li, H.~Zhang, Y.~Zhuang, Z.~Chen, Y.~Huang, Y.~Wang, Y.~Xu,
  D.~Zhuo, E.~P. Xing \emph{et~al.}, ``Alpa: Automating inter-and
  $\{$Intra-Operator$\}$ parallelism for distributed deep learning,'' in
  \emph{16th USENIX Symposium on Operating Systems Design and Implementation
  (OSDI 22)}, 2022, pp. 559--578.

\bibitem{chowdhery2022palm}
A.~Chowdhery, S.~Narang, J.~Devlin, M.~Bosma, G.~Mishra, A.~Roberts, P.~Barham,
  H.~W. Chung, C.~Sutton, S.~Gehrmann \emph{et~al.}, ``Palm: Scaling language
  modeling with pathways,'' \emph{arXiv preprint arXiv:2204.02311}, 2022.

\bibitem{chen2023frugalgpt}
L.~Chen, M.~Zaharia, and J.~Zou, ``Frugalgpt: How to use large language models
  while reducing cost and improving performance,'' \emph{arXiv preprint
  arXiv:2305.05176}, 2023.

\bibitem{chen2024data}
D.~Chen, Y.~Zhuang, S.~Zhang, J.~Liu, S.~Dong, and S.~Tang, ``Data shunt:
  Collaboration of small and large models for lower costs and better
  performance,'' in \emph{Proceedings of the AAAI Conference on Artificial
  Intelligence}, vol.~38, no.~10, 2024, pp. 11\,249--11\,257.

\bibitem{french1999catastrophic}
R.~M. French, ``Catastrophic forgetting in connectionist networks,''
  \emph{Trends in cognitive sciences}, vol.~3, no.~4, pp. 128--135, 1999.

\bibitem{gou2021knowledge}
J.~Gou, B.~Yu, S.~J. Maybank, and D.~Tao, ``Knowledge distillation: A survey,''
  \emph{International Journal of Computer Vision}, vol. 129, pp. 1789--1819,
  2021.

\bibitem{hershey2007approximating}
J.~R. Hershey and P.~A. Olsen, ``Approximating the kullback leibler divergence
  between gaussian mixture models,'' in \emph{2007 IEEE International
  Conference on Acoustics, Speech and Signal Processing-ICASSP'07},
  vol.~4.\hskip 1em plus 0.5em minus 0.4em\relax IEEE, 2007, pp. IV--317.

\bibitem{he2016ups}
R.~He and J.~McAuley, ``Ups and downs: Modeling the visual evolution of fashion
  trends with one-class collaborative filtering,'' in \emph{proceedings of the
  25th international conference on world wide web}, 2016, pp. 507--517.

\bibitem{mcauley2015image}
J.~McAuley, C.~Targett, Q.~Shi, and A.~Van Den~Hengel, ``Image-based
  recommendations on styles and substitutes,'' in \emph{Proceedings of the 38th
  international ACM SIGIR conference on research and development in information
  retrieval}, 2015, pp. 43--52.

\bibitem{kim2014convolutional}
Y.~Kim, ``Convolutional neural networks for sentence classification,''
  \emph{arXiv preprint arXiv:1408.5882}, 2014.

\bibitem{wang2016attention}
Y.~Wang, M.~Huang, X.~Zhu, and L.~Zhao, ``Attention-based lstm for aspect-level
  sentiment classification,'' in \emph{Proceedings of the 2016 conference on
  empirical methods in natural language processing}, 2016, pp. 606--615.

\bibitem{devlin2018BERT}
J.~Devlin, M.-W. Chang, K.~Lee, and K.~Toutanova, ``Bert: Pre-training of deep
  bidirectional transformers for language understanding,'' \emph{arXiv preprint
  arXiv:1810.04805}, 2018.

\bibitem{ynie2020chaosnli}
Y.~Nie, X.~Zhou, and M.~Bansal, ``What can we learn from collective human
  opinions on natural language inference data?'' in \emph{Proceedings of the
  2020 Conference on Empirical Methods in Natural Language Processing
  (EMNLP)}.\hskip 1em plus 0.5em minus 0.4em\relax Association for
  Computational Linguistics, 2020.

\bibitem{lewis2019bart}
M.~Lewis, Y.~Liu, N.~Goyal, M.~Ghazvininejad, A.~Mohamed, O.~Levy, V.~Stoyanov,
  and L.~Zettlemoyer, ``Bart: Denoising sequence-to-sequence pre-training for
  natural language generation, translation, and comprehension,'' \emph{arXiv
  preprint arXiv:1910.13461}, 2019.

\bibitem{yang2019xlnet}
Z.~Yang, Z.~Dai, Y.~Yang, J.~Carbonell, R.~R. Salakhutdinov, and Q.~V. Le,
  ``Xlnet: Generalized autoregressive pretraining for language understanding,''
  \emph{Advances in neural information processing systems}, vol.~32, 2019.

\bibitem{lan2019albert}
Z.~Lan, M.~Chen, S.~Goodman, K.~Gimpel, P.~Sharma, and R.~Soricut, ``Albert: A
  lite bert for self-supervised learning of language representations,''
  \emph{arXiv preprint arXiv:1909.11942}, 2019.

\bibitem{sanh2019distilbert}
V.~Sanh, L.~Debut, J.~Chaumond, and T.~Wolf, ``Distilbert, a distilled version
  of bert: smaller, faster, cheaper and lighter,'' \emph{arXiv preprint
  arXiv:1910.01108}, 2019.

\bibitem{zhou2018deep}
Y.~Zhou, Q.~Hu, and Y.~Wang, ``Deep super-class learning for long-tail
  distributed image classification,'' \emph{Pattern Recognition}, vol.~80, pp.
  118--128, 2018.

\bibitem{li2022trustworthy}
B.~Li, Z.~Han, H.~Li, H.~Fu, and C.~Zhang, ``Trustworthy long-tailed
  classification,'' in \emph{Proceedings of the IEEE/CVF Conference on Computer
  Vision and Pattern Recognition}, 2022, pp. 6970--6979.

\bibitem{he2016deep}
K.~He, X.~Zhang, S.~Ren, and J.~Sun, ``Deep residual learning for image
  recognition,'' in \emph{Proceedings of the IEEE conference on computer vision
  and pattern recognition}, 2016, pp. 770--778.

\bibitem{stefanini2022show}
M.~Stefanini, M.~Cornia, L.~Baraldi, S.~Cascianelli, G.~Fiameni, and
  R.~Cucchiara, ``From show to tell: A survey on deep learning-based image
  captioning,'' \emph{IEEE transactions on pattern analysis and machine
  intelligence}, vol.~45, no.~1, pp. 539--559, 2022.

\bibitem{xu2015show}
K.~Xu, J.~Ba, R.~Kiros, K.~Cho, A.~Courville, R.~Salakhudinov, R.~Zemel, and
  Y.~Bengio, ``Show, attend and tell: Neural image caption generation with
  visual attention,'' in \emph{International conference on machine
  learning}.\hskip 1em plus 0.5em minus 0.4em\relax PMLR, 2015, pp. 2048--2057.

\bibitem{lin2014microsoft}
T.-Y. Lin, M.~Maire, S.~Belongie, J.~Hays, P.~Perona, D.~Ramanan,
  P.~Doll{\'a}r, and C.~L. Zitnick, ``Microsoft coco: Common objects in
  context,'' in \emph{Computer Vision--ECCV 2014: 13th European Conference,
  Zurich, Switzerland, September 6-12, 2014, Proceedings, Part V 13}.\hskip 1em
  plus 0.5em minus 0.4em\relax Springer, 2014, pp. 740--755.

\bibitem{papineni2002bleu}
K.~Papineni, S.~Roukos, T.~Ward, and W.-J. Zhu, ``Bleu: a method for automatic
  evaluation of machine translation,'' in \emph{Proceedings of the 40th annual
  meeting of the Association for Computational Linguistics}, 2002, pp.
  311--318.

\bibitem{hu2021lora}
E.~J. Hu, P.~Wallis, Z.~Allen-Zhu, Y.~Li, S.~Wang, L.~Wang, W.~Chen
  \emph{et~al.}, ``Lora: Low-rank adaptation of large language models,'' in
  \emph{International Conference on Learning Representations}, 2021.

\bibitem{meta2024introducing}
A.~Meta, ``Introducing meta llama 3: The most capable openly available llm to
  date,'' \emph{Meta AI.}, 2024.

\end{thebibliography}

%




\end{document}